\documentclass[conference]{IEEEtran}
\IEEEoverridecommandlockouts
\usepackage{lineno} 
\usepackage{cite}
\usepackage{amsmath,amssymb,amsfonts}
\usepackage{amsmath}
\usepackage{amssymb}
\usepackage{graphicx}
\usepackage{textcomp}
\usepackage{xcolor}
\usepackage{algorithm}
\usepackage{algpseudocode}
\usepackage{booktabs} 
\usepackage{hyperref}
\usepackage{multirow}

\makeatletter
\newcommand{\linebreakand}{%
  \end{@IEEEauthorhalign}
  \hfill\mbox{}\par
  \mbox{}\hfill\begin{@IEEEauthorhalign}
}
\makeatother

\def\BibTeX{{\rm B\kern-.05em{\sc i\kern-.025em b}\kern-.08em
    T\kern-.1667em\lower.7ex\hbox{E}\kern-.125emX}}
\begin{document}

\title{GeMix: Conditional GAN-Based Mixup for Improved Medical Image Augmentation}

\author{\IEEEauthorblockN{Hugo Carlesso$^*$\thanks{$^*$Equal contribution.}}
\IEEEauthorblockA{
\textit{IRIT, UMR5505 CNRS}\\
\textit{Universit\'{e} de Toulouse}\\
Toulouse, France \\
hugo.carlesso@irit.fr}
\and
\IEEEauthorblockN{Maria Eliza Patulea$^*$}
\IEEEauthorblockA{\textit{Department of Computer Science} \\
\textit{University of Bucharest}\\
Bucharest, Romania \\
maria.patulea@s.unibuc.ro}
\and
\IEEEauthorblockN{Moncef Garouani}
\IEEEauthorblockA{\textit{IRIT, UMR5505 CNRS} \\
\textit{Universit\'{e} Toulouse Capitole}\\
Toulouse, France \\
moncef.garouani@irit.fr}
\linebreakand
\IEEEauthorblockN{Radu Tudor Ionescu}
\IEEEauthorblockA{\textit{Department of Computer Science} \\
\textit{University of Bucharest}\\
Bucharest, Romania \\
raducu.ionescu@gmail.com}
\and
\IEEEauthorblockN{Josiane Mothe}
\IEEEauthorblockA{\textit{IRIT, UMR5505 CNRS} \\
\textit{Universit\'{e} de Toulouse}\\
Toulouse, France \\
josiane.mothe@irit.fr}
}

\maketitle

\begin{abstract}
Mixup has become a popular augmentation strategy for image classification, yet its naive pixel-wise interpolation often produces unrealistic images that can hinder learning, particularly in high-stakes medical applications. We propose GeMix, a two-stage framework that replaces heuristic blending with a learned, label-aware interpolation powered by class-conditional GANs. First, a StyleGAN2-ADA generator is trained on the target dataset. During augmentation, we sample two label vectors from Dirichlet priors biased toward different classes and blend them via a Beta-distributed coefficient. Then, we condition the generator on this soft label to synthesize visually coherent images that lie along a continuous class manifold.
We benchmark GeMix on the large-scale COVIDx-CT-3 dataset using three backbones (ResNet-50, ResNet-101, EfficientNet-B0). When combined with real data, our method increases macro-F1 over traditional mixup for all backbones, reducing the false negative rate for COVID-19 detection. GeMix is thus a drop-in replacement for pixel-space mixup, delivering stronger regularization and greater semantic fidelity, without disrupting existing training pipelines. We publicly release our code at \url{https://github.com/hugocarlesso/GeMix} to foster reproducibility and further research. 
\end{abstract}


\section{Introduction}
\label{sec:intro}

Deep neural networks have achieved remarkable success in image classification tasks, yet their performance often depends on large labeled datasets and effective regularization techniques. Mixup augmentation~\cite{zhang2018mixup} aims to address both issues. Mixup is a simple yet powerful data augmentation method, wherein pairs of training examples and their corresponding labels are linearly interpolated to create synthetic examples. 

While mixup enhances generalization and robustness, it suffers from a fundamental limitation: the pixel-wise interpolation between two distinct images often yields visually unrealistic and semantically ambiguous results, especially when the source images belong to dissimilar classes. This effect may be especially harmful in safety-critical domains, such as medical imaging, where faint texture cues may carry diagnostic meaning. Indeed, these synthetic samples may confuse the model or provide misleading training signals, especially in complex datasets where visual realism and semantic consistency are crucial for learning meaningful representations. 

In this work, we hypothesize that adding realistic and semantically consistent synthetic images during training can lead to improved classification performance. To this end, we  revisit mixup through the lens of conditional generative adversarial networks (cGANs) \cite{mirza2014conditionalgenerativeadversarialnets}. 

\begin{figure*}[!htbp]
  \centering
  \includegraphics[width=0.8\textwidth]{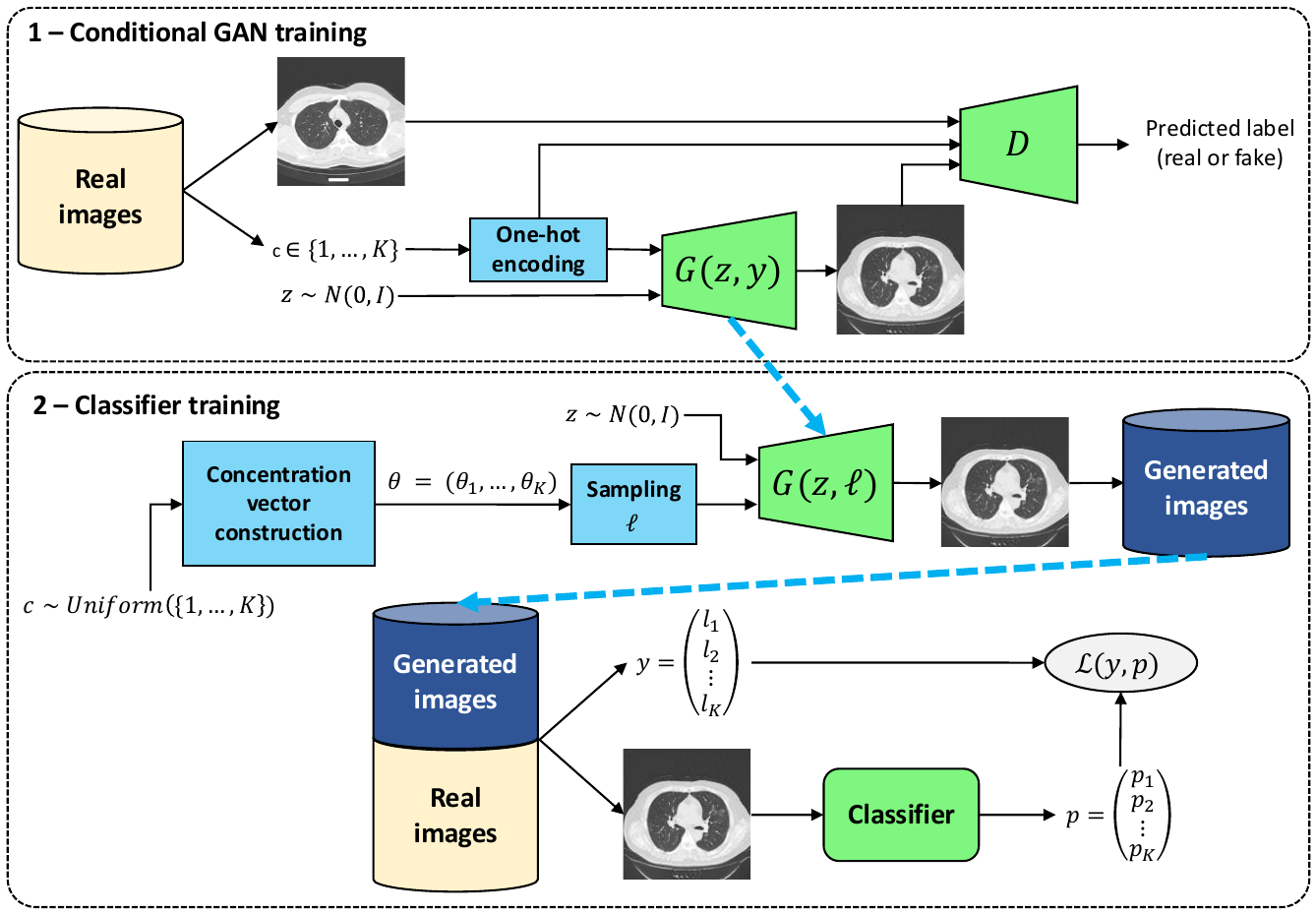}
  \caption{The proposed GeMix pipeline. In stage 1, a conditional GAN is trained on real images using one-hot encoded labels. In stage 2, images are generated with the trained conditional generator using soft labels sampled from a Dirichlet distribution biased toward a specific class. Generated images are stacked with real images and further used as augmented data to train a classifier.} 
  \label{fig:full_pipeline}
\end{figure*}

We introduce GeMix, a two-stage procedure that first trains a cGAN on real images and then synthesizes novel samples by soft-label interpolation, as illustrated in Figure \ref{fig:full_pipeline}. More specifically, we randomly pick a target class, draw a soft-label vector from a Dirichlet distribution biased toward that class, sample a Gaussian noise vector, and feed both into the generator to create a new image.
The generator, conditioned on mixed class labels, produces images that lie between classes on a learned data manifold, overcoming the unrealistic transitions of pixel-level mixup, while retaining controllable label information. By combining the flexibility of GAN-based image synthesis with the regularization benefits of label mixing,  GeMix provides a principled alternative to traditional mixup. 

We evaluate GeMix on the large-scale 
COVIDx-CT-3 benchmark, comparing three backbone families (ResNet-50 \cite{he_deep_2016}, ResNet-101 \cite{he_deep_2016}, EfficientNet-B0 \cite{tan2019efficientnet}) under different augmentation regimes. For all backbones our method yields consistent gains in terms of macro-F1 over traditional mixup, while also reducing the false negative rate for COVID-19 detection. We also analyze confusion matrices as well as the visual coherence of generated samples, which are key indicators in high-stakes domains, such as healthcare.

Our contribution is fourfold:
\begin{itemize}
    \item  We introduce GeMix, a label-conditioned GAN-based augmentation strategy that replaces heuristic pixel blending with learned data-driven mixing.
    \item We employ soft-label Dirichlet sampling, a mechanism that yields a continuous label manifold and unifies intra- and inter-class augmentation under a single probabilistic scheme.
    \item We conduct a rigorous evaluation to benchmark GeMix on a large multi-center medical dataset using multiple architectures.
    \item To foster reproducibility and future research, we publicly release our code at \url{https://github.com/hugocarlesso/GeMix}.
\end{itemize}

\section{Related Work}

Data augmentation in deep learning is a technique where new training data is generated from existing data to improve model performance, being  particularly useful when the original dataset is small or imbalanced \cite{shorten_khoshgoftaar_2019, wang2024comprehensive}, or the model is not robust to image transformations \cite{sandru2022feature}. 

In medical applications, traditional geometric and photometric transformations (e.g.~rotations, crops, intensity adjustments, noise) have been shown to reduce overfitting and boost generalization~\cite{litjens_survey_2017}, but they do not synthesize novel anatomical patterns or rare pathologies. Methods based on mixing or region-removal address this challenge by generating hybrid examples. 

A popular augmentation method is \textit{mixup}\cite{zhang2018mixup}, which creates new training examples through linear interpolation of pairs of existing examples and their corresponding labels. 
Formally, given two labeled examples, $(x_i, y_i)$ and $(x_j, y_j)$, mixup generates a new synthetic labeled sample $(\tilde{x}, \tilde{y})$ as follows:
\begin{equation}
\tilde{x} = \lambda x_i + (1 - \lambda)x_j, \quad \tilde{y} = \lambda y_i + (1 - \lambda)y_j,
\end{equation}
where $\lambda \in [0, 1]$ is sampled from a Beta distribution, $\texttt{Beta}(\alpha, \alpha)$ for some hyperparameter $\alpha > 0$.

When applied to medical imaging, the pixel-wise blending of two distinct anatomical structures can produce visually unrealistic or clinically implausible samples. A few variants have been introduced, but they fail to address this particular issue. CutMix pastes patches between images with label weighting by area~\cite{Yun19}, performing a spatial region-based blending to enhance local feature learning. Manifold mixup~\cite{verma2019manifold} extends the interpolation to hidden feature spaces. Cutout occludes random regions to encourage global context reliance~\cite{devries_improved_2017}. 
AugMix~\cite{hendrycksaugmix} composes randomized chains of simple augmentations (e.g. rotation, scale, flip, etc.) and linearly combines them with the original image. This improves calibration and robustness. None of the above augmentation methods care about obtaining realistic images. Yet, there are a few attempts that consider this issue. Anatomy-guided variants can either use class activation maps to guide mixing, e.g.~SnapMix~\cite{huang2021snapmix}, or preserve organ masks during pasting to maintain precise boundaries, e.g.~KeepMix and KeepMask~\cite{liu2023mixing}. 

Generative models, notably GANs, have been employed to produce realistic medical scans~\cite{ristea2023cytran}. Following this trend, we hypothesize that more anatomically plausible images can be obtained by automatic image generation. Our approach harnesses GAN-based generation to answer this issue.

\section{GeMix Augmentation}
\label{Approach}

\subsection{Background: Generative Adversarial Networks}

Generative Adversarial Networks, introduced by Goodfellow et al.~\cite{goodfellow2020generative}, are a class of unsupervised generative models that learn to produce realistic data samples through a competitive training process between two neural networks, a generator $G$ and a discriminator $D$. Starting from a noise vector $z$, the generator aims to produce synthetic data samples that resemble the training data as closely as possible. In contrast, the discriminator attempts to distinguish between real and generated samples. These two networks are trained simultaneously in a mini-max game, where the generator tries to fool the discriminator and the discriminator improves its ability to detect fakes. Formally, the objective can be expressed as follows:
\begin{equation}
\min_G \max_D \mathbb{E}_{x \sim p_{\text{data}}} [\log D(x)] + \mathbb{E}_{z \sim p_z} [\log(1 - D(G(z)))].
\end{equation}
Over time, the generator is supposed to learn to approximate the true data distribution. Conditional GANs~\cite{mirza2014conditionalgenerativeadversarialnets} extend the original GAN framework by incorporating additional input, such as class labels or semantic attributes. Because of this conditioning, the generator is able to produce data corresponding to specific categories, which is especially useful in scenarios such as class-conditional image generation. Since class labels are used during training, the generative framework becomes supervised.

\subsection{Overview of GeMix}

The proposed generative augmentation framework is designed to enhance classification performance by synthesizing realistic and semantically meaningful samples (see  Figure~\ref{fig:full_pipeline}). The approach consists of two primary stages: (i) training of a conditional GAN on the original labeled dataset, and  (ii) employing the trained generator to produce augmented data through a novel soft-label mixup mechanism.

\subsection{Conditional GAN Training} A conditional GAN is first trained on the labeled dataset to model the conditional distribution of images given class information. The generator component of the cGAN is conditioned on class-specific label vectors and optimized to produce realistic samples that reflect the semantic structure of each class. Once trained, the generator serves as the foundation for generating synthetic data in the subsequent augmentation phase.

\begin{algorithm}[t]
    \caption{GeMix Augmentation}
    \label{alg_gmix}
\begin{algorithmic}[1]
\Require 
\Statex \quad $N$ \hspace{1.9em} Number of images to generate
\Statex \quad $a_{=}$ \hspace{1.7em} Dirichlet concentration at dominant class
\Statex \quad $a_{\neq}$ \hspace{1.7em} Dirichlet concentration at other classes
\Statex \quad $K$ \hspace{1.9em} Total number of classes
\Statex \quad ${G}$ \hspace{2.1em} Generator network mapping $(z,\ell)\to x$
\For{$i=1$ \textbf{to} $N$}
    \State Sample $z\sim\mathcal{N}(0,I)$
    \State Sample class $c\sim \texttt{Uniform}(\{1,\dots,K\})$
    \For{$j=1$ \textbf{to} $K$}
        \State $\theta_j \leftarrow \begin{cases}
        a_{=} & \text{if } j = c,\\
        a_{\neq} & \text{otherwise}
        \end{cases}$
    \EndFor
\State Sample soft label $\ell \sim \texttt{Dirichlet}(\theta)$
\State Generate image $x \leftarrow {G}(z,\ell)$
\EndFor
\Statex
\noindent
\textbf{Example parameters:} $N=30000;\ a_{\neq}=1;\ a_{=}=2.$
\end{algorithmic}
\end{algorithm}

\subsection{GAN-Based Mixup Augmentation}
\label{GAN_mixup}
Building on the trained cGAN, we implement a generative data augmentation strategy aimed at increasing sample diversity and improving generalization in downstream tasks. The generator is employed to synthesize new samples conditioned on interpolated soft labels, enabling the creation of images that embody nuanced class mixtures. 

This augmentation mechanism is formally described in Algorithm \ref{alg_gmix}. The generator ${G}$ is conditioned on a soft label vector $\ell \in [0,1]^K$, with $K$ being the number of classes. Given a latent variable (noise) $z \sim \mathcal{N}(0, I)$ as input, it produces a synthetic sample $x = G(z, \ell)$.

Although the conditional GAN is trained solely on real samples with discrete class labels, we hypothesize that it learns a sufficiently smooth class-conditional latent space, such that interpolating the conditioning label vector leads to semantically meaningful transitions. This approach is inspired by prior work in conditional generative modeling~\cite{odena2017conditional, miyato2018cgans}, which demonstrated that even without explicit supervision for intermediate classes, interpolating soft labels can yield coherent outputs that blend characteristics of the involved categories. In our context, we observe that using Dirichlet-sampled soft labels enables the generator to produce anatomically consistent images that reflect plausible mixtures of clinical features. While the generator does not explicitly observe between-class samples during training, the interpolation property is central to the utility of GeMix for augmentation.
Here, the soft label of each sample is drawn by first selecting a single class index:
\begin{equation}
c \sim \texttt{Uniform}(\{1,\dots,K\}).
\end{equation}
We refer to the selected class $c$ as the ``dominant'' class, and all other classes as ``non-dominant'' classes. A
concentration vector \(\theta\in\mathbb{R}^K\) is then constructed:
\begin{equation}
\theta = (\theta_1, \dots, \theta_K),
\quad
\theta_j =
\begin{cases}
a_{=}, & j = c, \\
a_{\neq}, & j \neq c,
\end{cases}
\end{equation}
where \(a_{=}>a_{\neq}>0\) control the amount of mass placed on the dominant
class relative to the others. In practice, we set $a_{=}=2$ and $a_{\neq}=1$,
which ensures that the resulting soft label retains a clear emphasis on the chosen class, while still
incorporating useful contributions from the others. 

We then sample $\ell \sim \texttt{Dirichlet}(\theta)$, and feed the pair \((z,\ell)\) into \({G}\) to generate the synthetic image \(x\).

To generalize the traditional mixup using only two classes per blended image with a Beta distribution to multiple classes per image, the Dirichlet distribution is a natural choice. Yielding probability vectors that sum to one, it allows fine-grained control over concentration around the dominant class, enabling soft-label interpolation.
By adjusting \(\theta\), we can smoothly interpolate between class prototypes, while guaranteeing valid mixture weights and encouraging diverse soft labels.

Repeating this process \(N\) times yields a set of augmented pairs
\(\{(x_i,\ell_i)\}_{i=1}^N\), which are appended to the original training set. The effectiveness
of the proposed augmentation framework is assessed through a series of experiments described in
Section \ref{experiments}.

\section{Experiments}
\label{experiments}
\subsection{Dataset}

To assess the benefits of the GeMix augmentation strategy, we conduct experiments on the COVIDx CT-3 dataset \cite{Gunraj2020,Gunraj2022}, a large-scale open-access benchmark for COVID-19 detection from chest CT scans. COVIDx CT-3 comprises a total of 431,205 CT slices from 6,068 patients across more than 17 countries, making it the largest and most diverse public chest CT dataset available for this task. The dataset encompasses three classes, namely COVID-19, community-acquired pneumonia (CAP), and normal. It includes geographic diversity as well as carefully curated labels obtained via expert annotation and model-based selection methods, with validation and test sets fully annotated by experts to ensure high-quality evaluation.

Three balanced subsets of the COVIDx CT-3 dataset are independently selected via uniform sampling. The first one is used to train the conditional GAN and consists of 10,000 images per class, comprising a total of 30,000 samples. The original CT
slices exhibit varying resolutions, most measuring 512$\times$512 pixels. A second equally-balanced set of 30,000 images is independently sampled for the classification task. This set of images is partitioned into 80\% for training and 20\% for validation. Finally, model performance is evaluated on an independent test set comprising 1,000 images per class.

\subsection{Implementation Details}

To generate medical images, we employ StyleGAN2-ADA \cite{9156570}, an advanced variant of StyleGAN2 optimized for data-limited scenarios. The implementation is sourced from the official NVIDIA repository \cite{nvlabs_2021} and used to train a conditional GAN on lung CT images, forming the basis of our GeMix augmentation framework. All images are resized to 128$\times$128 pixels to ensure
compatibility with the StyleGAN2 training pipeline. 

Class labels are encoded as one-hot vectors. The StyleGAN2 model is trained on mini-batches of 32 samples, without image mirroring. StyleGAN2 is trained on Google Colab Pro with NVIDIA A100 GPUs, with checkpoints and data persisted to Google Drive. 
After training, the generator is employed to synthesize augmented samples using the proposed soft-label mixup strategy, as detailed in Algorithm~\ref{alg_gmix}. 

All classifiers are trained on an NVIDIA RTX 2000 Ada Generation GPU for 5 epochs with a batch size of 64  and a learning rate fixed to $10^{-4}$. All models are implemented in Python 3.11 using PyTorch 2.6 (CUDA 12.4).

\subsection{Baselines}

In standard mixup, new samples are generated by convexly combining inputs and labels. We generalize this framework to a multi-class setting by first randomly drawing one image from each of the \(K\) categories, and then sampling a soft-label vector \(\ell\) from a Dirichlet distribution biased toward a randomly chosen pivot class (its concentration set to 2, the others to 1), as described in Section \ref{GAN_mixup}. Let \(x_1,\dots,x_K\) be the selected images. We blend them pixel-wise according to:
\begin{equation}
x_{\mathrm{mix}} \;=\; \sum_{j=1}^K \ell_j\,x_j,
\qquad
y_{\mathrm{mix}} \;=\; \ell = (\ell_1,\dots,\ell_K),
\label{eq:traditional_mixup}
\end{equation}
where each \(\ell_j\in[0,1]\) denotes both the fraction of class \(j\) in the soft label
\(y_{\mathrm{mix}}\) and the relative contribution of image \(x_j\) to the mixed sample
\(x_{\mathrm{mix}}\).
Repeating this procedure \(N\) times yields a richly diversified set of synthetic training examples.

To distinguish the above version from standard mixup~\cite{zhang2018mixup}, we refer to the generalized version as \emph{multi-image
mixup} (MMixup). Note that MMixup can be seen as an ablated version of our GeMix, where there is no generator involved. We compare GeMix with both mixup and MMixup.

\subsection{Training Setups}

To assess the effectiveness of the proposed data augmentation strategy for COVID-19 CT image classification, we train several models under the following training setups: 
\begin{itemize}
    \item \textbf{Real:} 24K original CT images;
    \item \textbf{Mixup:} 24K images interpolated via traditional mixup;
    \item \textbf{MMixup:} 24K images interpolated via multi-image mixup;
    \item \textbf{GeMix:} 24K synthetic images generated via GeMix;
    \item \textbf{Real+Mixup:} 24K original images and 30K images obtained via traditional mixup;
    \item \textbf{Real+MMixup:} 24K original images and 30K images obtained via multi-image mixup;
    \item \textbf{Real+GeMix:} 24K original images and 30K images generated via GeMix;
    \item \textbf{Real+MMixup+GeMix:} 24K original images, 24K images obtained via multi-image mixup, and 24K images generated via GeMix.
\end{itemize}

The core comparison is between combinations of real images and augmented images either via Mixup, MMixup or GeMix, namely Real+Mixup, Real+MMixup and~Real+GeMix. These setups are directly comparable as they use the same number of mixed images. We also report results using only augmented images as input, showcasing the effect of not using real (unmodified) images during training. Finally, in the last setup, we compile a training set that comprises real images and images generated with both MMixup and GeMix.

MMixup corresponds to an ablation of our proposed GeMix method. It allows us to directly observe the improvement due to the use of generated images by GANs versus the generalization of mixup to the multi-class setting.

Each training setup is applied on three state-of-the-art deep learning architectures: ResNet-50, ResNet-101 and EfficientNet-B0. All models are pretrained on ImageNet \cite{deng_imagenet_2009}, enabling them to leverage transfer learning for improved performance on medical imaging tasks.

\begin{table}[t]
\caption{Classification results on the COVIDx CT-3 dataset using ResNet-50, ResNet-101 and EfficientNet-B0 architectures. Best results of each sub-group for a given architecture are in \textbf{bold}. Best results for a given model are \underline{underlined}.}
\renewcommand{\arraystretch}{1.2}
\centering
  \begin{tabular}{clccc}
    
    Model & Setup & P & R & F1\\
    \hline
    \hline
     \multirow{8}{*}{\textit{\rotatebox[origin=c]{90}{ResNet-50}}}& Real & 0.894 & 0.887 & 0.888 \\
    \cline{2-5}
    & Mixup \cite{zhang2018mixup} & \textbf{0.597}&\textbf{0.615} & \textbf{0.598}\\
    & MMixup & 0.444 & 0.592 & 0.479 \\
    & GeMix (ours) & 0.545 & 0.550 & 0.530 \\
    \cline{2-5}
    & Real+Mixup \cite{zhang2018mixup} & 0.902& 0.902& 0.902\\
    & Real+MMixup & 0.886 & 0.883 & 0.884 \\
    & Real+GeMix (ours) & \underline{\textbf{0.914}} & \underline{\textbf{0.910}} & \underline{\textbf{0.911}} \\
    \cline{2-5}
    & Real+MMixup+GeMix (ours) & 0.850 & 0.846 & 0.845 \\    
    \hline
    
    \multirow{8}{*}{\textit{\rotatebox[origin=c]{90}{ResNet-101}}}  & Real & {0.924} & {0.918} & {0.919} \\
    \cline{2-5}
    & Mixup \cite{zhang2018mixup} & 0.538&\textbf{0.564} & \textbf{0.538}\\
          & MMixup & 0.457 & 0.557 & 0.482 \\
    & GeMix (ours) & \textbf{0.548}& 0.553 & 0.527 \\
    \cline{2-5}
     & Real+Mixup \cite{zhang2018mixup} & 0.908 & 0.898 & 0.899\\
     & Real+MMixup & {0.918} & {0.914} & 0.914 \\
     & Real+GeMix (ours) & \textbf{\underline{0.924}} & \textbf{\underline{0.920}} & \textbf{\underline{0.921}} \\
    \cline{2-5}
    & Real+MMixup+GeMix (ours) & 0.914 & 0.912 & 0.913 \\
     \hline
     
    \multirow{8}{*}{\textit{\rotatebox[origin=c]{90}{EfficientNet-B0}}} & Real & 0.905 & 0.901 & 0.902 \\
    \cline{2-5}
    & Mixup \cite{zhang2018mixup} & \textbf{0.531} &0.553 & \textbf{0.538}\\
    & MMixup & 0.511 & \textbf{0.554} & 0.513 \\
    & GeMix (ours) & 0.500 & 0.498 & 0.466 \\
    \cline{2-5}
    & Real+Mixup \cite{zhang2018mixup} & 0.900 & 0.897 & 0.898 \\
    & Real+MMixup & 0.900 & 0.895 & 0.896 \\
    & Real+GeMix (ours) & \textbf{0.907} & \textbf{0.901} & \textbf{0.902} \\
    \cline{2-5}
    & Real+MMixup+GeMix (ours) & \underline{0.910} & \underline{0.908} & \underline{0.908} \\
    \hline
    \end{tabular}
\label{table:classificationresults}
\end{table}

\begin{figure}[t]
  \centering
  \setlength{\tabcolsep}{3pt}
  \begin{tabular}{ccc}
  Mixup&MMixup&\textbf{GeMix}\\
    \includegraphics[width=0.15\textwidth]{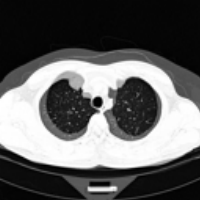} &
    \includegraphics[width=0.15\textwidth]{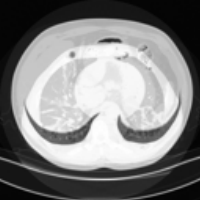} &
    \includegraphics[width=0.15\textwidth]{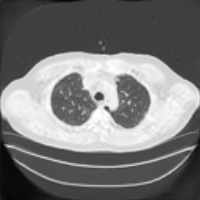} \\[2pt]
    \includegraphics[width=0.15\textwidth]{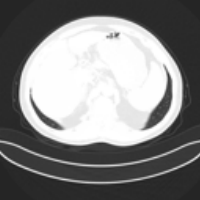} &
    \includegraphics[width=0.15\textwidth]{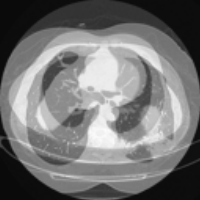} &
    \includegraphics[width=0.15\textwidth]{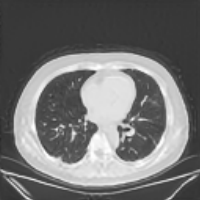} \\[2pt]
    \includegraphics[width=0.15\textwidth]{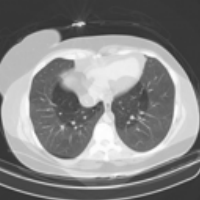} &
    \includegraphics[width=0.15\textwidth]{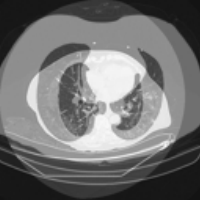} &
    \includegraphics[width=0.15\textwidth]{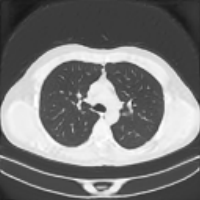} \\
  \end{tabular}
  \caption{Visual comparison of data augmentation strategies. Mixup \cite{zhang2018mixup} (first column) and MMixup (second column) produce images that blend information in a fashion that is not anatomically plausible. In contrast, GeMix (third column) produces images that are anatomically valid, since the interpolation is applied to the class-conditioning input of a GAN.}
  \label{fig:mixup_comparison}
\end{figure}

\subsection{Quantitative Results}

We report the macro-averaged precision (P), recall (R) and F1-score for each training scenario and architecture in Table~\ref{table:classificationresults}. In our balanced class distribution setting, the recall is equivalent to the accuracy rate.

When using only augmented images as input, Mixup is almost always better than GeMix. However, the performance levels of all augmentation strategies are far below the performance obtained by using only real images, indicating that using only augmented data is not sufficient to obtain robust models.

When combining augmented samples with real data, the ranking among augmentation strategies changes. Across all models, combining real data with GeMix consistently outperforms the combination with traditional mixup (Real+Mixup) or multi-image mixup (Real+MMixup). For instance, for ResNet-101, the Real+GeMix setting achieves the best performance across all metrics (recall is 0.920 and F1 is 0.921), surpassing both Real+Mixup and Real+MMixup configurations. 

We observe that the Real+Mixup and Real+MMixup settings are consistently below the Real setup, indicating that performing mixup in the original image space tends to degrade performance in medical imaging. This suggests that the use of images that are not anatomically plausible can degrade performance. In contrast, performing the mixing in the class space and using the mixed classes to condition a GAN leads to performance improvements.

Combining real images with multiple augmentation strategies can be seen as a straightforward way to boost performance. However, this setup, denoted as Real+MMixup+GeMix, exhibits performance drops for ResNet-50 and ResNet-101. The only model for which the Real+MMixup+GeMix setup works is EfficientNet-B0.

Overall, the results demonstrate that augmenting real CT data with GAN-generated images using soft-label mixup improves classification performance. Moreover, combining multiple augmentation strategies (as in Real+MMixup+GeMix) can sometimes provide additional benefits, particularly for EfficientNet-B0.

\begin{figure}[t]
  \centering
  \includegraphics[width=0.4\textwidth]{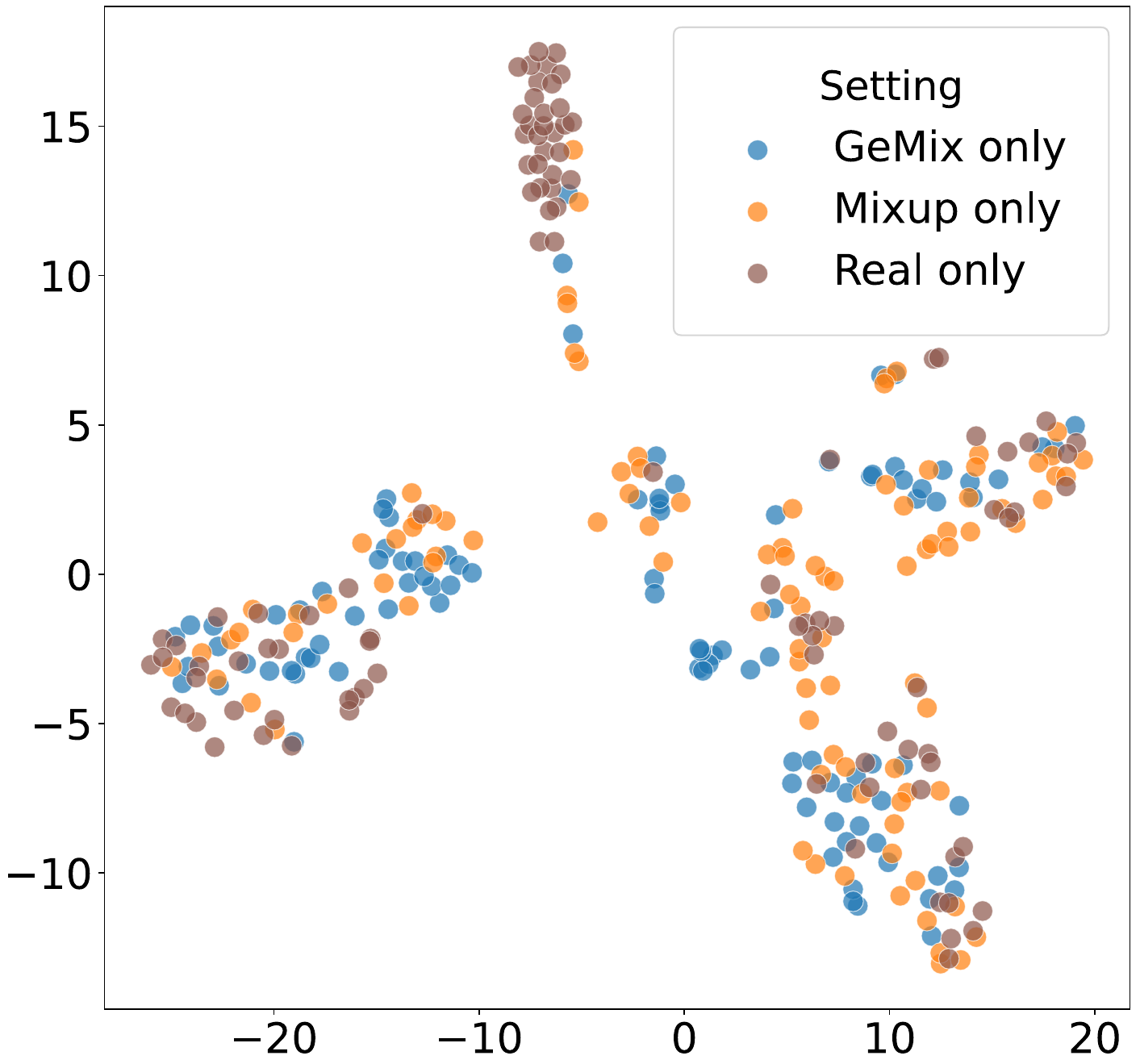}
  \caption{t-SNE visualization of 100 encoded samples per augmentation settings using ResNet-50 trained with real data only. Best viewed in color.}
  \label{fig:tse_plot}
\end{figure}

\begin{figure}[t!]
\centering
\includegraphics[width=0.24\textwidth]{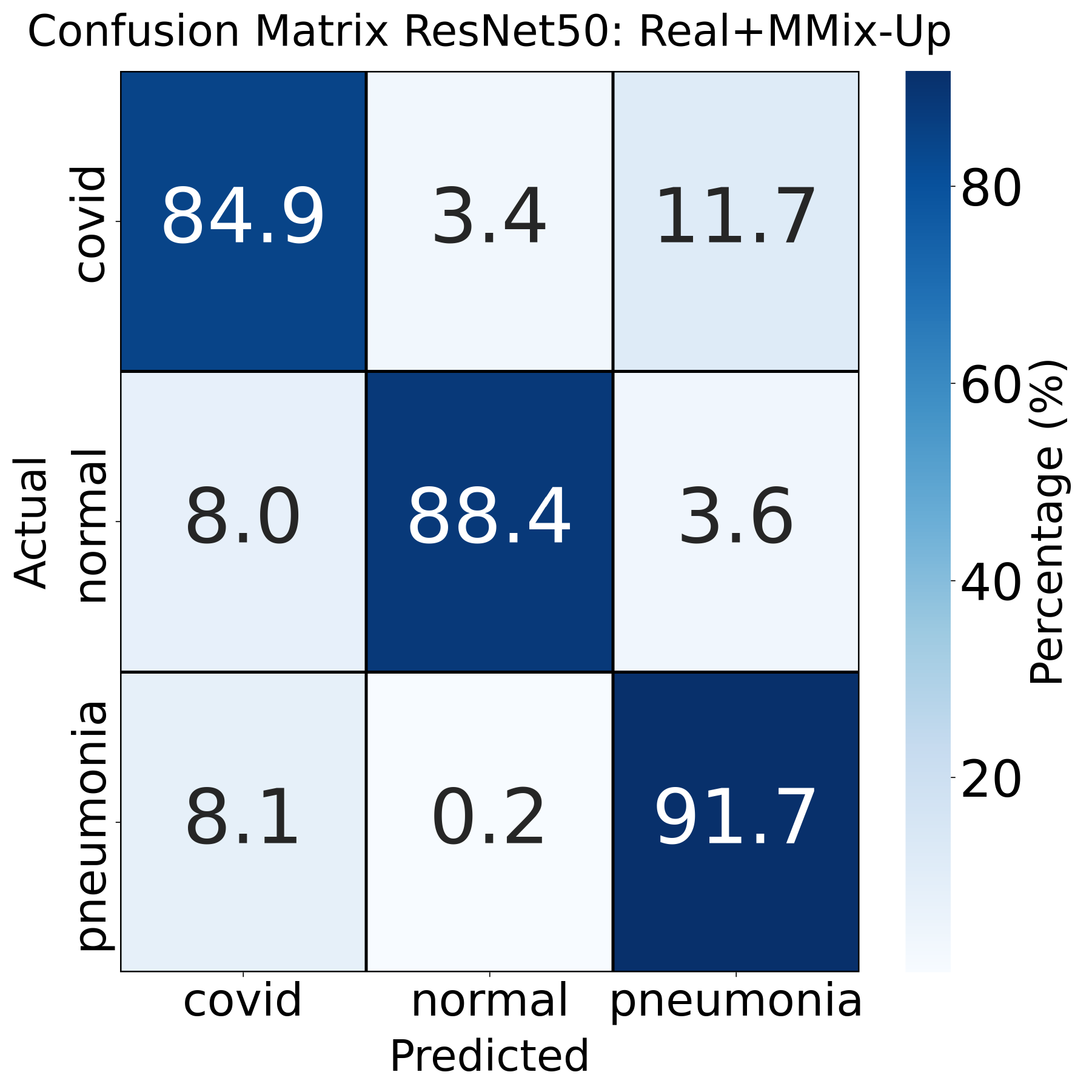}\hfill
\includegraphics[width=0.24\textwidth]{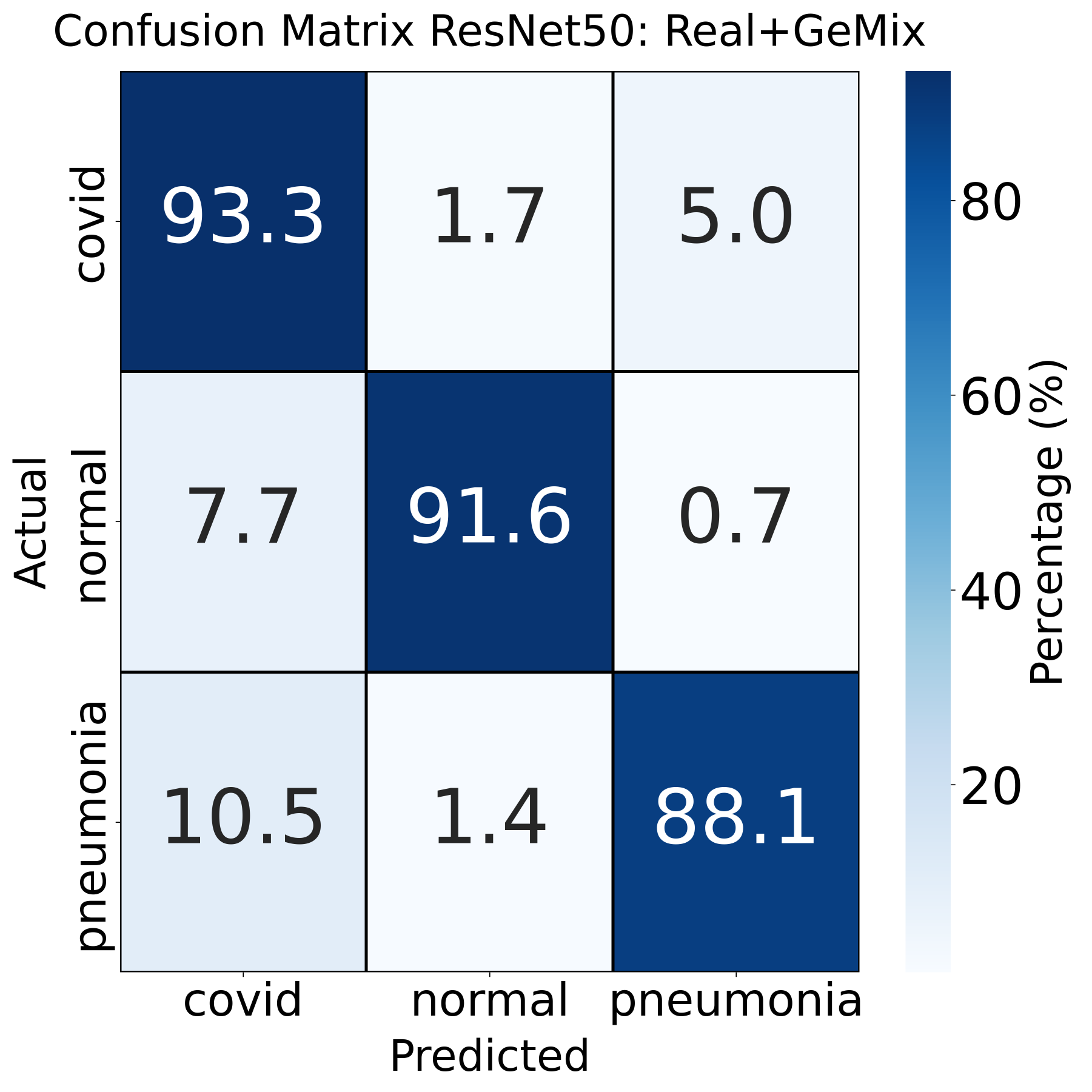}
\includegraphics[width=0.24\textwidth]{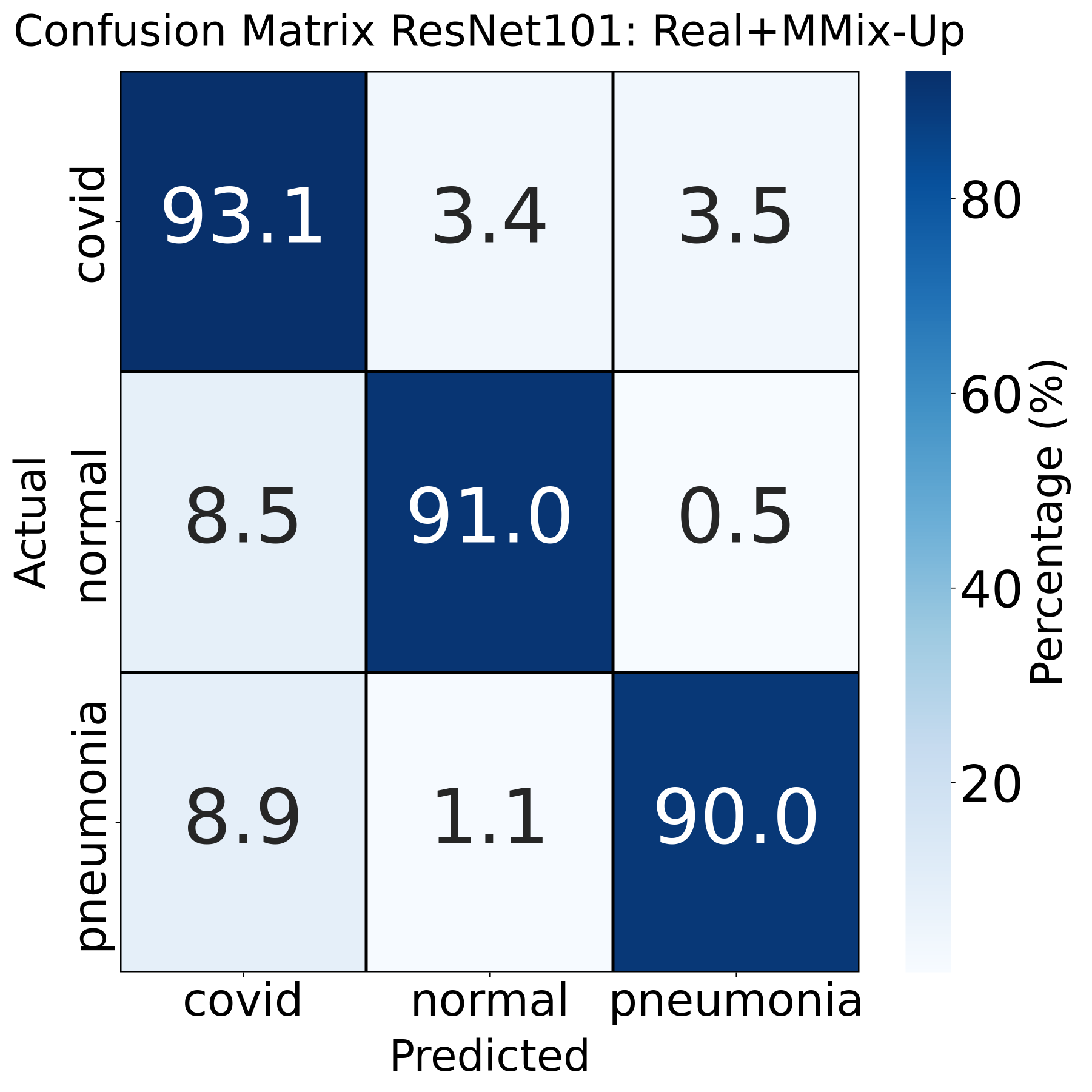}\hfill
\includegraphics[width=0.24\textwidth]{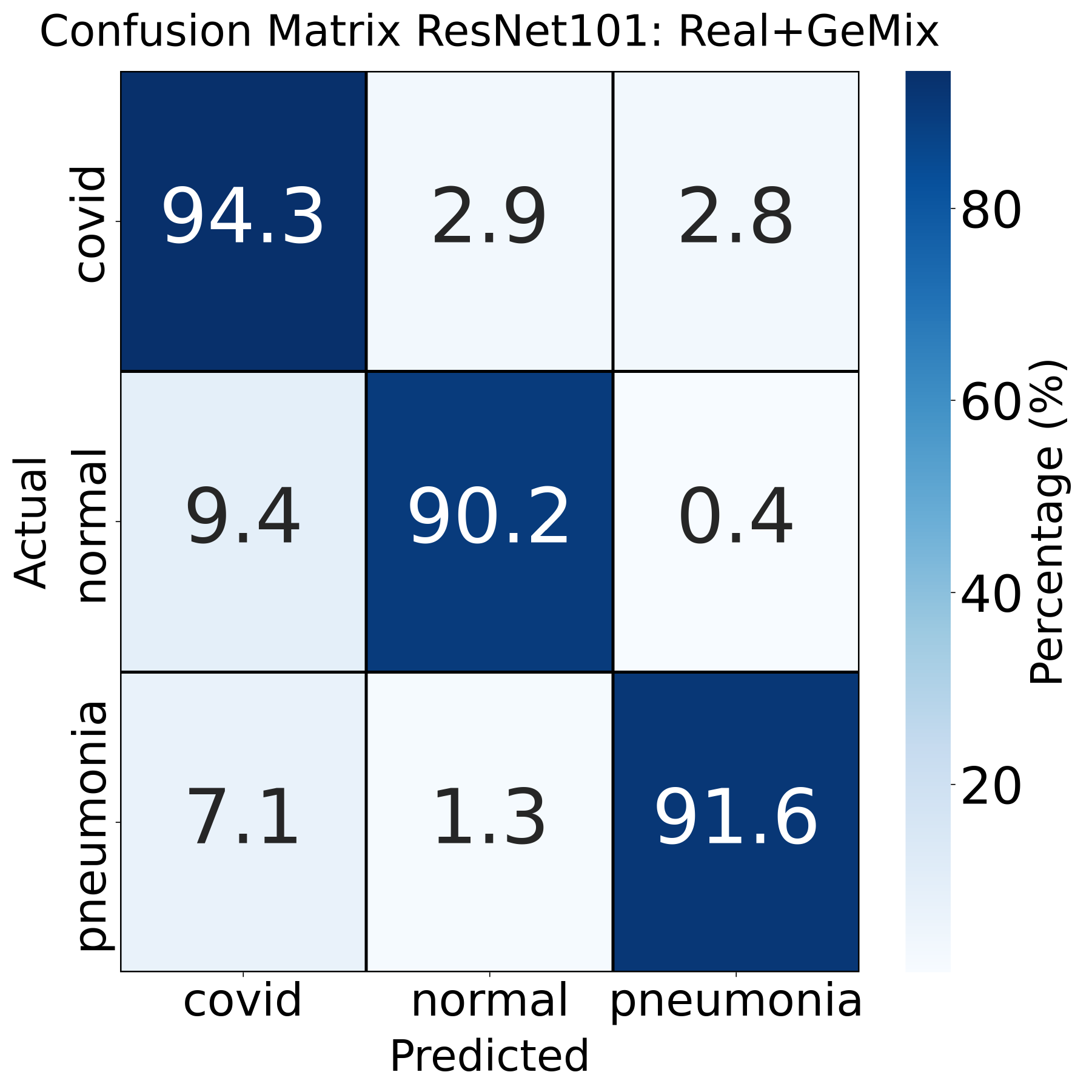}
\caption{Confusion matrices of Real+MMixup (left column) vs.~Real+GeMix (right column) across ResNet-50 (top row) and ResNet-101 (bottom row) architectures.}
\label{fig:confusion_matrices}
\end{figure}

\subsection{Qualitative Analysis}
\label{Qualitative}
\subsubsection{GeMix leads to more realistic images}
One of our hypotheses is that GeMix is supposed to lead to more realistic images than traditional mixup or MMixup. Figure~\ref{fig:mixup_comparison} illustrates a few examples produced by mixup (first column), MMixup (second column) and GeMix (third column). We observe that GeMix produces more anatomically coherent images, whereas the interpolation in pixel-space performed by mixup and MMixup leads to images that are not anatomically valid.

\subsubsection{GeMix expands the data distribution}

In Figure \ref{fig:tse_plot}, we use t-distributed Stochastic Neighbor Embedding (t-SNE) \cite{maaten2008visualizing} to visualize the ResNet-50 latent features. We compare embeddings of real data samples, mixup data samples and GeMix data samples, respectively. We observe that the variety of real data samples is lower than that of augmented samples. In terms of data distribution expansion via augmentation, we observe a slight edge in favor of GeMix. This could explain why GeMix leads to higher relative improvements than mixup.

\subsubsection{GeMix provides a lower false negative rate}
To better understand the effects of MMixup and GeMix data augmentation strategies, we conduct a qualitative error analysis based on the confusion matrices of each configuration. The confusion matrices for ResNet-50 and ResNet-101 are shown in Figure \ref{fig:confusion_matrices}.
When comparing the Real+GeMix and Real+MMixup settings across architectures, we observe that our conditional GAN-based augmentation consistently increases the number of true positives (TP), while reducing false negatives (FN), for both ResNet models. This indicates improved sensitivity and better detection of COVID-19 positive cases when GAN-generated samples are used during training. For ResNet-101, for instance, the Real+GeMix configuration shows a higher TP count than Real+MMixup, suggesting that the synthetic GAN-based images help the classification model to generalize better to positive cases, without increasing false positives (FP).

\section{Limitations}

While GeMix yields encouraging results and provides a novel perspective on GAN-based data augmentation for medical imaging, some limitations warrant consideration. Our experiments are exclusively conducted on the COVIDx-CT-3 dataset, which consists of chest CT scans with anatomically consistent content. Although this choice enables focused evaluation, it does not confirm that our findings are also applicable to datasets with broader anatomical or modality diversity. Additionally, although our method is grounded in the assumption that realism in generated images enhances classifier performance, results observed with synthetic data alone suggest that this relationship may be more nuanced and context-dependent. The conditional GAN employed in GeMix is trained using only single-class images. While the empirical results demonstrate its capacity to produce meaningful interpolations, a more explicit modeling of inter-class transitions could further enhance this capability. Moreover, the resolution of input images was set to \(128 \times 128\) for computational efficiency, which may constrain the representation of fine-grained anatomical features. While the hyperparameters for Dirichlet sampling are selected empirically and remain fixed throughout the experiments, further exploring their influence could offer additional insights into the flexibility and robustness of the proposed method. Finally, although qualitative improvements in sample realism are evident, expert clinical validation would provide stronger support for the medical plausibility of generated images. Extending GeMix to higher-resolution settings, more varied datasets, and more comprehensive evaluation protocols constitutes a promising direction for future research.

\section{Conclusion}

This paper introduced GeMix, an extension of mixup based on conditional GANs, which replaces pixel-level interpolation with learned image-label blending. 
A StyleGAN2-ADA generator trained on COVIDx-CT-3 was conditioned on Dirichlet-sampled soft labels to synthesize realistic and semantically aligned CT slices. When the synthetic images were combined with real data, various models (ResNet-50, ResNet-101, EfficientNet-B0) achieved consistent gains in macro-F1 over traditional mixup, lowering the false negative rate for COVID-19 detection. These results confirm that label-aware generative mixing can deliver stronger regularization than heuristic pixel blending.

In future work, we plan to apply GeMix to additional domains, including natural image benchmarks, in order to determine how well label-aware generative mixing generalizes across various types of images and class-imbalance profiles. We will also analyze the benefit of applying GeMix to Vision Transformers \cite{dosovitskiy_beyer_kolesnikov_weissenborn_zhai_unterthiner_dehghani_minderer_heigold_gelly_et_al_2021}.

\section*{Acknowledgment}

This work was supported by a grant of the Ministry of Research, Innovation and Digitization, CCCDI - UEFISCDI, project number PN-IV-P7-7.1-PED-2024-1856, within PNCDI IV. This project has received financial support from the CNRS through the MITI interdisciplinary programs.

\bibliography{CBMI}
\bibliographystyle{IEEEtran}
\end{document}